\documentclass{edm_template} 
\usepackage{times}
\usepackage{amsmath,epsfig,cite} 
\usepackage{amsfonts}
\usepackage{amssymb}
\usepackage{mathrsfs}
\usepackage{subfigure}
\usepackage[algo2e]{algorithm2e}
\usepackage{amsmath}
\usepackage{paralist}
\usepackage{booktabs}
\usepackage{subfigure}
\usepackage{color}
\usepackage{url}
\DeclareGraphicsExtensions{.pdf,.jpeg,.png,.epbibdesk}
\usepackage{epstopdf}
\usepackage{cite}
\usepackage[T1]{fontenc}
\usepackage{graphicx}
\usepackage{multirow}
\usepackage{algpseudocode}

\usepackage{floatrow}
\newfloatcommand{capbtabbox}{table}[][\FBwidth]

\usepackage{yhmath}

\usepackage{amssymb}
\usepackage{amsfonts}
\usepackage{mathrsfs}
\usepackage{xspace}
\usepackage{bm}
\usepackage{upgreek}

\newcommand{\safemath}[2]{\newcommand{#1}{\ensuremath{#2}\xspace}}

\safemath{\bma}{\mathbf{a}}
\safemath{\bmb}{\mathbf{b}}
\safemath{\bmc}{\mathbf{c}}
\safemath{\bmd}{\mathbf{d}}
\safemath{\bme}{\mathbf{e}}
\safemath{\bmf}{\mathbf{f}}
\safemath{\bmg}{\mathbf{g}}
\safemath{\bmh}{\mathbf{h}}
\safemath{\bmi}{\mathbf{i}}
\safemath{\bmj}{\mathbf{j}}
\safemath{\bmk}{\mathbf{k}}
\safemath{\bml}{\mathbf{l}}
\safemath{\bmm}{\mathbf{m}}
\safemath{\bmn}{\mathbf{n}}
\safemath{\bmo}{\mathbf{o}}
\safemath{\bmp}{\mathbf{p}}
\safemath{\bmq}{\mathbf{q}}
\safemath{\bmr}{\mathbf{r}}
\safemath{\bms}{\mathbf{s}}
\safemath{\bmt}{\mathbf{t}}
\safemath{\bmu}{\mathbf{u}}
\safemath{\bmv}{\mathbf{v}}
\safemath{\bmw}{\mathbf{w}}
\safemath{\bmx}{\mathbf{x}}
\safemath{\bmy}{\mathbf{y}}
\safemath{\bmz}{\mathbf{z}}
\safemath{\bmzero}{\mathbf{0}}
\safemath{\bmone}{\mathbf{1}}

\bmdefine{\biad}{a}
\bmdefine{\bibd}{b}
\bmdefine{\bicd}{c}
\bmdefine{\bidd}{d}
\bmdefine{\bied}{e}
\bmdefine{\bifd}{f}
\bmdefine{\bigd}{g}
\bmdefine{\bihd}{h}
\bmdefine{\biid}{i}
\bmdefine{\bijd}{j}
\bmdefine{\bikd}{k}
\bmdefine{\bild}{l}
\bmdefine{\bimd}{m}
\bmdefine{\bind}{n}
\bmdefine{\biod}{o}
\bmdefine{\bipd}{p}
\bmdefine{\biqd}{q}
\bmdefine{\bird}{r}
\bmdefine{\bisd}{s}
\bmdefine{\bitd}{t}
\bmdefine{\biud}{u}
\bmdefine{\bivd}{v}
\bmdefine{\biwd}{w}
\bmdefine{\bixd}{x}
\bmdefine{\biyd}{y}
\bmdefine{\bizd}{z}

\bmdefine{\bixid}{\xi}
\bmdefine{\bilambdad}{\lambda}
\bmdefine{\bimud}{\mu}
\bmdefine{\bithetad}{\theta}
\bmdefine{\biphid}{\phi}
\bmdefine{\bideltad}{\delta}

\safemath{\bmia}{\biad}
\safemath{\bmib}{\bibd}
\safemath{\bmic}{\bicd}
\safemath{\bmid}{\bidd}
\safemath{\bmie}{\bied}
\safemath{\bmif}{\bifd}
\safemath{\bmig}{\bigd}
\safemath{\bmih}{\bihd}
\safemath{\bmii}{\biid}
\safemath{\bmij}{\bijd}
\safemath{\bmik}{\bikd}
\safemath{\bmil}{\bild}
\safemath{\bmim}{\bimd}
\safemath{\bmin}{\bind}
\safemath{\bmio}{\biod}
\safemath{\bmip}{\bipd}
\safemath{\bmiq}{\biqd}
\safemath{\bmir}{\bird}
\safemath{\bmis}{\bisd}
\safemath{\bmit}{\bitd}
\safemath{\bmiu}{\biud}
\safemath{\bmiv}{\bivd}
\safemath{\bmiw}{\biwd}
\safemath{\bmix}{\bixd}
\safemath{\bmiy}{\biyd}
\safemath{\bmiz}{\bizd}

\safemath{\bmxi}{\bixid}
\safemath{\bmlambda}{\bilambdad}
\safemath{\bmmu}{\bimud}
\safemath{\bmtheta}{\bithetad}
\safemath{\bmphi}{\biphid}
\safemath{\bmdelta}{\bideltad}

\safemath{\bA}{\mathbf{A}}
\safemath{\bB}{\mathbf{B}}
\safemath{\bC}{\mathbf{C}}
\safemath{\bD}{\mathbf{D}}
\safemath{\bE}{\mathbf{E}}
\safemath{\bF}{\mathbf{F}}
\safemath{\bG}{\mathbf{G}}
\safemath{\bH}{\mathbf{H}}
\safemath{\bI}{\mathbf{I}}
\safemath{\bJ}{\mathbf{J}}
\safemath{\bK}{\mathbf{K}}
\safemath{\bL}{\mathbf{L}}
\safemath{\bM}{\mathbf{M}}
\safemath{\bN}{\mathbf{N}}
\safemath{\bO}{\mathbf{O}}
\safemath{\bP}{\mathbf{P}}
\safemath{\bQ}{\mathbf{Q}}
\safemath{\bR}{\mathbf{R}}
\safemath{\bS}{\mathbf{S}}
\safemath{\bT}{\mathbf{T}}
\safemath{\bU}{\mathbf{U}}
\safemath{\bV}{\mathbf{V}}
\safemath{\bW}{\mathbf{W}}
\safemath{\bX}{\mathbf{X}}
\safemath{\bY}{\mathbf{Y}}
\safemath{\bZ}{\mathbf{Z}}

\safemath{\bZero}{\mathbf{0}}
\safemath{\bOne}{\mathbf{1}}
\safemath{\bDelta}{\mathbf{\Delta}}
\safemath{\bLambda}{\mathbf{\UpLambda}}
\safemath{\bPhi}{\mathbf{\Upphi}}
\safemath{\bSigma}{\mathbf{\Upsigma}}
\safemath{\bOmega}{\mathbf{\Upomega}}
\safemath{\bTheta}{\mathbf{\Uptheta}}

\bmdefine{\biAd}{A}
\bmdefine{\biBd}{B}
\bmdefine{\biCd}{C}
\bmdefine{\biDd}{D}
\bmdefine{\biEd}{E}
\bmdefine{\biFd}{F}
\bmdefine{\biGd}{G}
\bmdefine{\biHd}{H}
\bmdefine{\biId}{I}
\bmdefine{\biJd}{J}
\bmdefine{\biKd}{K}
\bmdefine{\biLd}{L}
\bmdefine{\biMd}{M}
\bmdefine{\biNd}{N}
\bmdefine{\biOd}{O}
\bmdefine{\biPd}{P}
\bmdefine{\biQd}{Q}
\bmdefine{\biRd}{R}
\bmdefine{\biSd}{S}
\bmdefine{\biTd}{T}
\bmdefine{\biUd}{U}
\bmdefine{\biVd}{V}
\bmdefine{\biWd}{W}
\bmdefine{\biXd}{X}
\bmdefine{\biYd}{Y}
\bmdefine{\biZd}{Z}

\bmdefine{\biDelta}{\Delta}
\bmdefine{\biLambda}{\Lambda}
\bmdefine{\biPhi}{\Phi}
\bmdefine{\biSigma}{\Sigma}
\bmdefine{\biOmega}{\Omega}
\bmdefine{\biTheta}{\Theta}

\safemath{\bimA}{\biAd}
\safemath{\bimB}{\biBd}
\safemath{\bimC}{\biCd}
\safemath{\bimD}{\biDd}
\safemath{\bimE}{\biEd}
\safemath{\bimF}{\biFd}
\safemath{\bimG}{\biGd}
\safemath{\bimH}{\biHd}
\safemath{\bimI}{\biId}
\safemath{\bimJ}{\biJd}
\safemath{\bimK}{\biKd}
\safemath{\bimL}{\biLd}
\safemath{\bimM}{\biMd}
\safemath{\bimN}{\biNd}
\safemath{\bimO}{\biOd}
\safemath{\bimP}{\biPd}
\safemath{\bimQ}{\biQd}
\safemath{\bimR}{\biRd}
\safemath{\bimS}{\biSd}
\safemath{\bimT}{\biTd}
\safemath{\bimU}{\biUd}
\safemath{\bimV}{\biVd}
\safemath{\bimW}{\biWd}
\safemath{\bimX}{\biXd}
\safemath{\bimY}{\biYd}
\safemath{\bimZ}{\biZd}

\safemath{\bimDelta}{\biDelta}
\safemath{\bimLambda}{\biLambda}
\safemath{\bimPhi}{\biPhi}
\safemath{\bimSigma}{\biSigma}
\safemath{\bimOmega}{\biOmega}
\safemath{\bimTheta}{\biTheta}

\safemath{\setA}{\mathcal{A}}
\safemath{\setB}{\mathcal{B}}
\safemath{\setC}{\mathcal{C}}
\safemath{\setD}{\mathcal{D}}
\safemath{\setE}{\mathcal{E}}
\safemath{\setF}{\mathcal{F}}
\safemath{\setG}{\mathcal{G}}
\safemath{\setH}{\mathcal{H}}
\safemath{\setI}{\mathcal{I}}
\safemath{\setJ}{\mathcal{J}}
\safemath{\setK}{\mathcal{K}}
\safemath{\setL}{\mathcal{L}}
\safemath{\setM}{\mathcal{M}}
\safemath{\setN}{\mathcal{N}}
\safemath{\setO}{\mathcal{O}}
\safemath{\setP}{\mathcal{P}}
\safemath{\setQ}{\mathcal{Q}}
\safemath{\setR}{\mathcal{R}}
\safemath{\setS}{\mathcal{S}}
\safemath{\setT}{\mathcal{T}}
\safemath{\setU}{\mathcal{U}}
\safemath{\setV}{\mathcal{V}}
\safemath{\setW}{\mathcal{W}}
\safemath{\setX}{\mathcal{X}}
\safemath{\setY}{\mathcal{Y}}
\safemath{\setZ}{\mathcal{Z}}
\safemath{\emptySet}{\varnothing}

\safemath{\colA}{\mathscr{A}}
\safemath{\colB}{\mathscr{B}}
\safemath{\colC}{\mathscr{C}}
\safemath{\colD}{\mathscr{D}}
\safemath{\colE}{\mathscr{E}}
\safemath{\colF}{\mathscr{F}}
\safemath{\colG}{\mathscr{G}}
\safemath{\colH}{\mathscr{H}}
\safemath{\colI}{\mathscr{I}}
\safemath{\colJ}{\mathscr{J}}
\safemath{\colK}{\mathscr{K}}
\safemath{\colL}{\mathscr{L}}
\safemath{\colM}{\mathscr{M}}
\safemath{\colN}{\mathscr{N}}
\safemath{\colO}{\mathscr{O}}
\safemath{\colP}{\mathscr{P}}
\safemath{\colQ}{\mathscr{Q}}
\safemath{\colR}{\mathscr{R}}
\safemath{\colS}{\mathscr{S}}
\safemath{\colT}{\mathscr{T}}
\safemath{\colU}{\mathscr{U}}
\safemath{\colV}{\mathscr{V}}
\safemath{\colW}{\mathscr{W}}
\safemath{\colX}{\mathscr{X}}
\safemath{\colY}{\mathscr{Y}}
\safemath{\colZ}{\mathscr{Z}}

\safemath{\opA}{\mathbb{A}}
\safemath{\opB}{\mathbb{B}}
\safemath{\opC}{\mathbb{C}}
\safemath{\opD}{\mathbb{D}}
\safemath{\opE}{\mathbb{E}}
\safemath{\opF}{\mathbb{F}}
\safemath{\opG}{\mathbb{G}}
\safemath{\opH}{\mathbb{H}}
\safemath{\opI}{\mathbb{I}}
\safemath{\opJ}{\mathbb{J}}
\safemath{\opK}{\mathbb{K}}
\safemath{\opL}{\mathbb{L}}
\safemath{\opM}{\mathbb{M}}
\safemath{\opN}{\mathbb{N}}
\safemath{\opO}{\mathbb{O}}
\safemath{\opP}{\mathbb{P}}
\safemath{\opQ}{\mathbb{Q}}
\safemath{\opR}{\mathbb{R}}
\safemath{\opS}{\mathbb{S}}
\safemath{\opT}{\mathbb{T}}
\safemath{\opU}{\mathbb{U}}
\safemath{\opV}{\mathbb{V}}
\safemath{\opW}{\mathbb{W}}
\safemath{\opX}{\mathbb{X}}
\safemath{\opY}{\mathbb{Y}}
\safemath{\opZ}{\mathbb{Z}}
\safemath{\opZero}{\mathbb{O}}
\safemath{\identityop}{\opI}

\safemath{\veca}{\bma}
\safemath{\vecb}{\bmb}
\safemath{\vecc}{\bmc}
\safemath{\vecd}{\bmd}
\safemath{\vece}{\bme}
\safemath{\vecf}{\bmf}
\safemath{\vecg}{\bmg}
\safemath{\vech}{\bmh}
\safemath{\veci}{\bmi}
\safemath{\vecj}{\bmj}
\safemath{\veck}{\bmk}
\safemath{\vecl}{\bml}
\safemath{\vecm}{\bmm}
\safemath{\vecn}{\bmn}
\safemath{\veco}{\bmo}
\safemath{\vecp}{\bmp}
\safemath{\vecq}{\bmq}
\safemath{\vecr}{\bmr}
\safemath{\vecs}{\bms}
\safemath{\vect}{\bmt}
\safemath{\vecu}{\bmu}
\safemath{\vecv}{\bmv}
\safemath{\vecw}{\bmw}
\safemath{\vecx}{\bmx}
\safemath{\vecy}{\bmy}
\safemath{\vecz}{\bmz}

\safemath{\veczero}{\bmzero}
\safemath{\vecone}{\bmone}
\safemath{\vecxi}{\bmxi}
\safemath{\veclambda}{\bmlambda}
\safemath{\vecmu}{\bmmu}
\safemath{\vectheta}{\bmtheta}
\safemath{\vecphi}{\bmphi}
\safemath{\vecdelta}{\bmdelta}

\safemath{\matA}{\bA}
\safemath{\matB}{\bB}
\safemath{\matC}{\bC}
\safemath{\matD}{\bD}
\safemath{\matE}{\bE}
\safemath{\matF}{\bF}
\safemath{\matG}{\bG}
\safemath{\matH}{\bH}
\safemath{\matI}{\bI}
\safemath{\matJ}{\bJ}
\safemath{\matK}{\bK}
\safemath{\matL}{\bL}
\safemath{\matM}{\bM}
\safemath{\matN}{\bN}
\safemath{\matO}{\bO}
\safemath{\matP}{\bP}
\safemath{\matQ}{\bQ}
\safemath{\matR}{\bR}
\safemath{\matS}{\bS}
\safemath{\matT}{\bT}
\safemath{\matU}{\bU}
\safemath{\matV}{\bV}
\safemath{\matW}{\bW}
\safemath{\matX}{\bX}
\safemath{\matY}{\bY}
\safemath{\matZ}{\bZ}
\safemath{\matzero}{\bmzero}

\safemath{\matDelta}{\bDelta}
\safemath{\matLambda}{\bLambda}
\safemath{\matPhi}{\bPhi}
\safemath{\matSigma}{\bSigma}
\safemath{\matOmega}{\bOmega}
\safemath{\matTheta}{\bTheta}

\safemath{\matidentity}{\matI}
\safemath{\matone}{\matO}

\safemath{\rnda}{A}
\safemath{\rndb}{B}
\safemath{\rndc}{C}
\safemath{\rndd}{D}
\safemath{\rnde}{E}
\safemath{\rndf}{F}
\safemath{\rndg}{G}
\safemath{\rndh}{H}
\safemath{\rndi}{I}
\safemath{\rndj}{J}
\safemath{\rndk}{K}
\safemath{\rndl}{L}
\safemath{\rndm}{M}
\safemath{\rndn}{N}
\safemath{\rndo}{O}
\safemath{\rndp}{P}
\safemath{\rndq}{Q}
\safemath{\rndr}{R}
\safemath{\rnds}{S}
\safemath{\rndt}{T}
\safemath{\rndu}{U}
\safemath{\rndv}{V}
\safemath{\rndw}{W}
\safemath{\rndx}{X}
\safemath{\rndy}{Y}
\safemath{\rndz}{Z}

\safemath{\rveca}{\bimA}
\safemath{\rvecb}{\bimB}
\safemath{\rvecc}{\bimC}
\safemath{\rvecd}{\bimD}
\safemath{\rvece}{\bimE}
\safemath{\rvecf}{\bimF}
\safemath{\rvecg}{\bimG}
\safemath{\rvech}{\bimH}
\safemath{\rveci}{\bimI}
\safemath{\rvecj}{\bimJ}
\safemath{\rveck}{\bimK}
\safemath{\rvecl}{\bimL}
\safemath{\rvecm}{\bimM}
\safemath{\rvecn}{\bimN}
\safemath{\rveco}{\bomO}
\safemath{\rvecp}{\bimP}
\safemath{\rvecq}{\bimQ}
\safemath{\rvecr}{\bimR}
\safemath{\rvecs}{\bimS}
\safemath{\rvect}{\bimT}
\safemath{\rvecu}{\bimU}
\safemath{\rvecv}{\bimV}
\safemath{\rvecw}{\bimW}
\safemath{\rvecx}{\bimX}
\safemath{\rvecy}{\bimY}
\safemath{\rvecz}{\bimZ}

\safemath{\rvecxi}{\bmxi}
\safemath{\rveclambda}{\bmlambda}
\safemath{\rvecmu}{\bmmu}
\safemath{\rvectheta}{\bmtheta}
\safemath{\rvecphi}{\bmphi}

\safemath{\rmatA}{\bimA}
\safemath{\rmatB}{\bimB}
\safemath{\rmatC}{\bimC}
\safemath{\rmatD}{\bimD}
\safemath{\rmatE}{\bimE}
\safemath{\rmatF}{\bimF}
\safemath{\rmatG}{\bimG}
\safemath{\rmatH}{\bimH}
\safemath{\rmatI}{\bimI}
\safemath{\rmatJ}{\bimJ}
\safemath{\rmatK}{\bimK}
\safemath{\rmatL}{\bimL}
\safemath{\rmatM}{\bimM}
\safemath{\rmatN}{\bimN}
\safemath{\rmatO}{\bimO}
\safemath{\rmatP}{\bimP}
\safemath{\rmatQ}{\bimQ}
\safemath{\rmatR}{\bimR}
\safemath{\rmatS}{\bimS}
\safemath{\rmatT}{\bimT}
\safemath{\rmatU}{\bimU}
\safemath{\rmatV}{\bimV}
\safemath{\rmatW}{\bimW}
\safemath{\rmatX}{\bimX}
\safemath{\rmatY}{\bimY}
\safemath{\rmatZ}{\bimZ}

\safemath{\rmatDelta}{\bimDelta}
\safemath{\rmatLambda}{\bimLambda}
\safemath{\rmatPhi}{\bimPhi}
\safemath{\rmatSigma}{\bimSigma}
\safemath{\rmatOmega}{\bimOmega}
\safemath{\rmatTheta}{\bimTheta}

\usepackage{amssymb}
\usepackage{amsfonts}
\usepackage{mathrsfs}
\usepackage{xspace}
\usepackage{bm}
\usepackage{fancyref}
\usepackage{textcomp}

\usepackage{multirow}
\usepackage{stmaryrd}

\newenvironment{textbmatrix}{	\setlength{\arraycolsep}{2.5pt}%
								\big[\begin{matrix}}{\end{matrix}\big]%
								\raisebox{0.08ex}{\vphantom{M}}}

\def\be{\begin{equation}}
\def\ee{\end{equation}}
\def\een{\nonumber \end{equation}}
\def\mat{\begin{bmatrix}}
\def\emat{\end{bmatrix}}
\def\btm{\begin{textbmatrix}}
\def\etm{\end{textbmatrix}}

\def\ba#1\ea{\begin{align}#1\end{align}}
\def\bas#1\eas{\begin{align*}#1\end{align*}}
\def\bs#1\es{\begin{split}#1\end{split}} 
\def\bg#1\eg{\begin{gather}#1\end{gather}}
\def\bml#1\eml{\begin{multline}#1\end{multline}}
\def\bi#1\ei{\begin{itemize}#1\end{itemize}}

\DeclareMathOperator{\rank}{rank}			

\safemath{\dirac}{\delta}					
\safemath{\krond}{\dirac}					

\safemath{\upto}{\uparrow}
\safemath{\downto}{\downarrow}
\safemath{\iu}{j}							
\safemath{\ev}{\lambda}						
\safemath{\hilseqspace}{l^{2}}				
\newcommand{\banachfunspace}[1]{\setL^{#1}}	
\safemath{\hilfunspace}{\banachfunspace{2}}	

\safemath{\SNR}{\text{\sc snr}} 				
\safemath{\No}{N_0}							
\safemath{\Es}{E_s}							
\safemath{\Eb}{E_b}							
\safemath{\EbNo}{\frac{\Eb}{\No}}
\safemath{\EsNo}{\frac{\Es}{\No}}

\DeclareMathOperator{\CHop}{\ensuremath{\opH}} 
\safemath{\tvir}{\rndh_{\CHop}}				
\safemath{\tvtf}{\rndl_{\CHop}}				
\safemath{\spf}{\rnds_{\CHop}}				
\safemath{\bff}{H_{\CHop}}					

\safemath{\ircf}{r_{h}}						
\safemath{\tftvcf}{r_{s}}					
\safemath{\tfcf}{r_{l}}						
\safemath{\bfcf}{r_{H}}						

\safemath{\tcorr}{c_h}						
\safemath{\scf}{c_{s}}						
\safemath{\tfcorr}{c_{l}}					
\safemath{\fcorr}{c_{H}}						

\safemath{\mi}{I}							
\safemath{\capacity}{C}						

\safemath{\normal}{\mathcal{N}}			
\safemath{\jpg}{\mathcal{CN}}			
\safemath{\mchain}{\leftrightarrow}		

\safemath{\dB}{\,\mathrm{dB}}
\safemath{\dBm}{\,\mathrm{dBm}}
\safemath{\Hz}{\,\mathrm{Hz}}
\safemath{\kHz}{\,\mathrm{kHz}}
\safemath{\MHz}{\,\mathrm{MHz}}
\safemath{\GHz}{\,\mathrm{GHz}}
\safemath{\s}{\,\mathrm{s}}
\safemath{\ms}{\,\mathrm{ms}}
\safemath{\mus}{\,\mathrm{\text{\textmu}s}}
\safemath{\ns}{\,\mathrm{ns}}
\safemath{\ps}{\,\mathrm{ps}}
\safemath{\meter}{\,\mathrm{m}}
\safemath{\mm}{\,\mathrm{mm}}
\safemath{\cm}{\,\mathrm{cm}}
\safemath{\m}{\,\mathrm{m}}
\safemath{\W}{\,\mathrm{W}}
\safemath{\mW}{\, \mathrm{mW}}
\safemath{\J}{\,\mathrm{J}}
\safemath{\K}{\,\mathrm{K}}
\safemath{\bit}{\,\mathrm{bit}}
\safemath{\nat}{\,\mathrm{nat}}

\safemath{\define}{\triangleq}			

\safemath{\equivalent}{\sim}
\safemath{\distas}{\sim}					
\safemath{\sdiff}{\Delta}				

\safemath{\reals}{\mathbb{R}}
\safemath{\positivereals}{\reals_{+}}
\safemath{\integers}{\mathbb{Z}}
\safemath{\posint}{\integers_{+}}
\safemath{\naturals}{\mathbb{N}}
\safemath{\posnaturals}{\naturals_{+}}
\safemath{\complexset}{\mathbb{C}}
\safemath{\rationals}{\mathbb{Q}}

\newcommand*{\fancyrefapplabelprefix}{app}		
\newcommand*{\fancyrefthmlabelprefix}{thm}		
\newcommand*{\fancyreflemlabelprefix}{lem}		
\newcommand*{\fancyrefcorlabelprefix}{cor}		
\newcommand*{\fancyrefdeflabelprefix}{def}		
\newcommand*{\fancyrefalglabelprefix}{alg}		
\newcommand*{\fancyrefproplabelprefix}{prop}		
\newcommand*{\fancyrefexmpllabelprefix}{exmpl}
\newcommand*{\fancyreftbllabelprefix}{tbl}
\frefformat{vario}{\fancyrefseclabelprefix}{Sec.~#1}
\frefformat{vario}{\fancyrefthmlabelprefix}{Thm.~#1}
\frefformat{vario}{\fancyreflemlabelprefix}{Lem.~#1}
\frefformat{vario}{\fancyrefcorlabelprefix}{Cor.~#1}
\frefformat{vario}{\fancyrefdeflabelprefix}{Def.~#1}
\frefformat{vario}{\fancyrefalglabelprefix}{Alg.~#1}
\frefformat{vario}{\fancyreffiglabelprefix}{Fig.~#1}
\frefformat{vario}{\fancyrefapplabelprefix}{App.~#1} 
\frefformat{vario}{\fancyrefeqlabelprefix}{(#1)}
\frefformat{vario}{\fancyrefproplabelprefix}{Prop.~#1}
\frefformat{vario}{\fancyrefexmpllabelprefix}{Ex.~#1}
\frefformat{vario}{\fancyreftbllabelprefix}{Tbl.~#1}

\safemath{\dictab}{[\,\dicta\,\,\dictb\,]}

\safemath{\ysig}{\bmy}
\safemath{\ysighat}{\hat{\ysig}}
\safemath{\ysigdim}{M}
\safemath{\xsig}{\bmx}
\safemath{\xsigdim}{N}
\safemath{\nx}{n_x}
\safemath{\zsig}{\bmz}
\safemath{\zsigdim}{\ysigdim}
\safemath{\rsig}{\bmr}
\safemath{\Adict}{\bA}
\safemath{\Adicttilde}{\widetilde{\Adict}}
\safemath{\Adictdim}{\outputdim\times\xsigdim}
\safemath{\avec}{\bma}
\safemath{\avectilde}{\tilde{\avec}}
\safemath{\Bdict}{\bB}
\safemath{\Bdicttilde}{\widetilde{\Bdict}}
\safemath{\Cdict}{\bC}
\safemath{\cvec}{\bmc}
\safemath{\Ddict}{\bD}
\safemath{\Ddictdim}{\ysigdim\times\xsigdim}
\safemath{\dvec}{\bmd}
\safemath{\Ddicttilde}{\widetilde{\bD}}
\safemath{\Bonb}{\bB}
\safemath{\bvec}{\bmb}
\safemath{\Bonbdim}{\ysigdim\times\ysigdim}
\safemath{\noise}{\bmn}
\safemath{\noisedim}{\ysigim}
\safemath{\err}{\bme}
\safemath{\errdim}{\ysigdim}
\safemath{\errset}{\setE}
\safemath{\nerr}{n_e}
\safemath{\delop}{\bP_\errset}
\safemath{\delopc}{\bP_{{\errset}^c}}

\safemath{\cplxi}{\imath}
\safemath{\cplxj}{\jmath}

\safemath{\dict}{\matD}
\safemath{\inputdim}{N}		
\safemath{\outputdim}{M}		
\safemath{\sparsity}{S}	
\safemath{\inputdimA}{{N_a}}	
\safemath{\inputdimB}{{N_b}}	
\safemath{\elemA}{{n_a}}	
\safemath{\elemB}{{n_b}}	
\safemath{\resA}{\matR_a}	
\safemath{\resB}{\matR_b}	
\safemath{\subD}{\matS} 
\safemath{\subA}{\matS_a} 
\safemath{\subB}{\matS_b} 
\safemath{\dicta}{\matA} 	
\safemath{\dictb}{\matB} 	
\safemath{\hollowS}{H}
\safemath{\hollowA}{H_a}
\safemath{\hollowB}{H_b}
\safemath{\cross}{Z}
\safemath{\coh}{\mu_d}			
\safemath{\coha}{\mu_a}			
\safemath{\cohb}{\mu_b}			
\safemath{\mubs}{\nu}	
\safemath{\cohm}{\mu_m} 
\safemath{\dictset}{\setD}	
\safemath{\dictsetp}{\dictset(\coh,\coha,\cohb)}	
\safemath{\dictsetgen}{\dictset_\text{gen}}
\safemath{\dictsetgenp}{\dictsetgen(\coh)}
\safemath{\dictsetonb}{\dictset_\text{onb}}
\safemath{\dictsetonbp}{\dictsetonb(\coh)}

\safemath{\leftside}{U}
\safemath{\rightsideA}{R_a}
\safemath{\rightsideB}{R_b}

\safemath{\indexS}{\setI_S} 

\safemath{\na}{n_a}			
\safemath{\nb}{n_b}			
\safemath{\coeffa}{p_i}	
\safemath{\coeffb}{q_j}	
\safemath{\seta}{\setP}		
\safemath{\setb}{\setQ}     
\safemath{\setw}{\setW}	
\safemath{\setz}{\setZ}	
\safemath{\cola}{\veca}		
\safemath{\colb}{\vecb}		
\safemath{\cold}{\vecd}		
\safemath{\inputvec}{\vecx} 	
\safemath{\error}{\vece}	
\safemath{\noiseout}{\vecz} 	
\safemath{\inputvecel}{x}
\safemath{\inputveca}{\vecx_a}
\safemath{\inputvecb}{\vecx_b}
\safemath{\outputvec}{\vecy}	
\safemath{\lambdamin}{\lambda_{\mathrm{min}}}

\safemath{\elltwo}{\ell_2}
\safemath{\ellone}{\ell_1}
\safemath{\ellzero}{\ell_0}
\safemath{\ellinf}{\ell_\infty}
\safemath{\licard}{Z(\coh,\coha,\cohb)}
\safemath{\xsol}{\hat{x}}
\safemath{\xbord}{x_b}		
\safemath{\xstat}{x_s}		
\safemath{\xstatLone}{\tilde{x}_s}
\safemath{\order}{\mathcal{O}} 
\safemath{\scales}{\Theta} 
\safemath{\ones}{\mathbf{1}} 
\safemath{\zeroes}{\mathbf{0}} 
\safemath{\thlone}{\kappa(\coh,\cohb)} 
\safemath{\constoneA}{\delta} 
\safemath{\constoneB}{\epsilon} 
\safemath{\nlarge}{L}				   
\safemath{\sumlarge}{S_\nlarge}
\safemath{\maxlarger}{P_\nlarge}	   
\safemath{\Pzero}{\textrm{P0}}	
\safemath{\Pone}{\textrm{P1}}
\safemath{\vecfir}{\vecw}			 
\safemath{\vecsec}{\vecz}
\safemath{\elvecfir}{w}              
\safemath{\elvecsec}{z}				 
\safemath{\nlargefir}{n}
\safemath{\normout}{\gamma}
\safemath{\auxfun}{h}
\safemath{\supp}{\textrm{supp}}

\safemath{\indexa}{\ell}
\safemath{\indexb}{r}
\safemath{\indexc}{i}
\safemath{\indexd}{j}

\safemath{\project}{P}

\begin{document}

\title{Quantized Matrix Completion for Personalized Learning}

\numberofauthors{3}
\author{
\alignauthor
Andrew S. Lan \\[0.25cm]
       \affaddr{Rice University}\\[0.1cm]
       \email{mr.lan@sparfa.com}
\alignauthor
Christoph Studer \\[0.25cm]
       \affaddr{Cornell University}\\[0.1cm]
       \email{studer@sparfa.com}
\alignauthor
Richard G. Baraniuk \\[0.25cm]
       \affaddr{Rice University}\\[0.1cm]
       \email{richb@sparfa.com}
       }



\maketitle

\begin{abstract}

\sloppy
The recently proposed SPARse Factor Analysis (SPARFA) framework for personalized learning performs \emph{factor analysis} on ordinal or binary-valued (e.g., correct/incorrect) graded learner responses to questions. 
The underlying factors are termed ``concepts'' (or knowledge components) and are used for \emph{learning analytics~(LA)}, the estimation of learner concept-knowledge profiles, and for \emph{content analytics (CA)}, the estimation of question--concept associations and question difficulties.
While SPARFA is a powerful tool for LA and CA, it requires a number of algorithm parameters (including the number of concepts), which are difficult to determine in practice.
In this paper, we propose \emph{SPARFA-Lite}, a convex optimization-based method for LA that builds on matrix completion, which only requires a \emph{single} algorithm parameter and enables us to automatically identify the required number of concepts. 
Using a variety of educational datasets, we demonstrate that SPARFA-Lite (i) achieves comparable performance in predicting unobserved learner responses to existing methods, including item response theory (IRT) and SPARFA, and (ii) is computationally more efficient.

\end{abstract}


\section{Introduction} \label{sec:intro}

Recent advances in machine learning propel the design of personalized learning systems (PLSs) that mine learner data (e.g., graded responses to tests or homework assignments) to automatically provide timely feedback to individual learners.
Such automated systems have the potential to revolutionize education by delivering a high-quality, personalized learning experience at large scale. 

\sloppy

\subsection{SPARse Factor Analysis (SPARFA)}

The recently proposed SPARse Factor Analysis (SPARFA) framework introduces models and machine learning algorithms for learning and content analytics \cite{sparfa,sparfatag}. 
\emph{Learning analytics} (LA) stands for the analysis of the knowledge of each learner, while \emph{content analytics} (CA) stands for the analysis of all learning resources, i.e., textbooks, lecture videos, questions, etc.
SPARFA analyzes binary-valued ($1$ for a correct answer and $0$ for an incorrect one) or quantized (ordinal-valued, e.g., partial credits) graded responses of $N$ learners to $Q$ questions, in the domain of a course/exam. 
The key assumption of SPARFA is that the learners' responses to questions are governed by a small number of $K$ ($K\ll N,Q$) latent factors, called ``concepts,''  which are also known as knowledge components \cite{kt}. 
SPARFA performs the joint estimation of (i) question--concept associations, (ii) learner concept knowledge profiles, and (iii) question difficulties, solely from binary-valued graded learner responses. 
Provided this analysis, SPARFA enables a PLS to provide automated feedback to learners on their individual concept knowledge and to course instructors on the content organization of the  analyzed course.

SPARFA, as well as other factor analysis methods, inevitably suffer from the lack of a principled and computationally efficient way to select the appropriate values of the algorithms' parameters, especially the number of latent concepts $K$. 
The choice of the number of concepts $K$ is important for two reasons: First, it affects the performance in predicting unobserved learner responses. Second, it determines the interpretability of the estimated concepts, which is key for a PLS to provide human-interpretable feedback to learners. 
Rule-based intelligent tutoring systems \cite{andes2005} rely on domain experts to manually pre-define the value of $K$. 
Such an approach turns out to be labor-intensive and is error prone, which prevents its use for applications in massive open online courses (MOOCs) \cite{mooc1}. 
SPARFA utilizes cross-validation to select $K$, as well as all other algorithm parameters \cite{sparfa}. 
Such an approach is computationally extensive as it requires multiple SPARFA runs to identify appropriate values for all algorithm parameters. 

\subsection{Contributions}

\sloppy

In this work, we propose SPARFA-Lite, a convex optimization-based LA algorithm that automatically selects the number of latent concepts $K$ by analyzing graded learner responses in the domain of a single course/assessment. 
SPARFA-Lite leverages recent results in quantized matrix completion \cite{qmc} to analyze quantized graded learner responses, which accounts for the fact that responses are often graded on an ordinal scale (partial credit). 
Since SPARFA-Lite only has a single algorithm parameter, it has low computational complexity as compared to existing methods such as IRT or conventional SPARFA. 
We demonstrate the effectiveness of SPARFA-Lite in (i) predicting unobserved learner responses and (ii) performing LA on a variety of real-world educational datasets. 

\fussy

\subsection{Related work}

\sloppy
Factor analysis has been used extensively to analyze graded learner response data \cite{stampersm,cohen}. 
While some factor analysis methods treat binary-valued graded learner responses as real numbers \cite{qmatrix,qprobit}, others use probabilistic models to achieve superior performance in predicting unobserved learner responses. These methods include the additive factor model (AFM) \cite{afm}, instructional factors analysis (IFA) \cite{ifm}, and learning factor analysis (LFA) \cite{lfa}, which all assume that the number of concepts $K$ to be known a priori.
Collaborative filtering IRT (CF-IRT) \cite{logitfa} and SPARFA \cite{sparfa} both use cross-validation to select~$K$, as well as all other tuning parameters, by identifying the best prediction performance on unobserved learner responses.
This approach is computationally extensive and does not scale to MOOC scale applications, where the dimension of the problem is large and immediate feedback is required.
The authors of \cite{predfact} proposed to select $K$ by applying an SVD to the binary-valued graded learner response matrix and examining the decay of its singular values, which is not an automated approach. 

Matrix completion (MC) aims to recover a low-rank matrix from incomplete, real-valued observations~\cite{mc1,svt}, and has been used extensively in collaborative filtering applications. 
More recently, \mbox{1-bit} MC \cite{1bitmc}, and its generalization, quantized MC \cite{qmc} have been proposed for the recovery of low-rank matrices from incomplete binary-valued and quantized (or ordinal) observations, respectively. Since the graded learner responses in educational scenarios are typically binary-valued or ordinal, we next investigate the  applicability of quantized MC~\cite{qmc}  to educational scenarios.


\section{SPARFA-Lite statistical model} 
\label{sec:model}

SPARFA-Lite aims at recovering the unknown, low-rank matrix~$\bZ$ that governs the learners' responses to questions, solely from quantized (ordinal) graded learner responses.
Suppose that we have $N$ learners answering $Q$ questions. Let the $Q \times N$ matrix $\bZ$ be the underlying low-rank matrix that we seek to recover. Let $Y_{i,j} \in \mathcal{O}$ denote the quantized observed graded response of the $j^\text{th}$ learner, with $j \in  \{1, \ldots, N\}$, to the $i^\text{th}$ question, with $i \in  \{1, \ldots, Q\}$. $\mathcal{O} = \{1, \ldots, P\} $ is a set of $P$ ordered labels. 
Inspired by~\cite{qmc}, we use the following model for the graded learner response $Y_{i,j}$:
\begin{align} \label{eq:qa}  
\begin{array}{l}
  Y_{i,j} = \mathcal{Q}(Z_{i,j}+\epsilon_{i,j}),  \; (i,j)\in\Omega_\text{obs}, \\[0.1cm]
  \epsilon_{i,j} \sim \textit{Logistic}\!\left(0, 1 \right).
  \end{array} 
\end{align}
Here, $\textit{Logistic}\!\left(0, 1 \right)$ represents the Logistic distribution with zero mean and unit scale \cite{tibsbook}. 
The set $\Omega_\text{obs}\subseteq\{1,\ldots,Q\}\times\{1,\ldots,N\}$ contains the indices associated to the observed learner responses~$Y_{i,j}$.
The function $\mathcal{Q}(\cdot)\!:\! \mathbb{R} \rightarrow \mathcal{O}$ represents a scalar quantizer, defined as
\begin{align*}
\mathcal{Q}(x) = 
p \quad \text{if }\, \omega_{p-1}< x \leq \omega_p,\,\,   p \in \setO,
\end{align*}
where $\{\omega_0, \ldots, \omega_P \}$ is a set of quantization bin boundaries, with $\omega_0 \leq\omega_1 \leq \cdots \leq \omega_{P-1} \leq \omega_P$. We will assume that the set of quantization bin boundaries $\{\omega_0, \ldots, \omega_P \}$ is known a priori. In situations where these bin boundaries are unknown, they can be estimated directly from data (see, e.g., \cite{qmc,sparfatag} for the details).
\sloppy

In terms of the likelihood of the observed graded learner responses~$Y_{i,j}$, the model in \fref{eq:qa} can be written equivalently as
\begin{align} \label{eq:qap}
 p(Y_{i,j}=p \mid Z_{i,j}) 
 & \! = \, \Phi \! \left(  \omega_p \! - \! Z_{i,j} \right) \! - \! \Phi \! \left( \omega_{p-1} \! - \! Z_{i,j}\right),
\end{align}
where $\Phi(x)= \frac{1}{1+e^{-x}}$ corresponds to the inverse logit link function. 
For this paper, we will be using only the inverse logit link function as it leads to algorithms with lower computational complexity comparing to the inverse probit link function \cite{qmc}. 

The goal of the SPARFA-Lite algorithm detailed next is to recover the unknown low-rank matrix $\bZ$ given the observed binary-valued graded learner responses $Y_{i,j}, (i,j)\in\Omega_\text{obs}$.

\section{The SPARFA-Lite algorithm} 
\label{sec:matrix}

To recover the low-rank matrix $\bZ$ from binary-valued graded learner responses, we minimize the negative log-likelihood of the observed graded learner responses $Y_{i,j}$, $(i,j)\in\Omega_\text{obs}$, subject to a low-rank promoting constraint on $\bZ$.
In particular, we seek to solve the following convex optimization problem:
\begin{align*}
(\text{P})
\left\{
\begin{array}{ll}
\underset{\bZ \in \mathbb{R}^{Q \times N}}{\text{minimize\,\,\,\,\,\,}} & f(\bZ) = \textstyle -\sum_{i,j: (i,j) \in \Omega_\text{obs}} \log p(Y_{i,j}|Z_{i,j}) \\[0.1cm]
 \text{subject to} & \!\! \| \bZ \| \leq \lambda.
 \end{array}\right.
\end{align*}
Here, the constraint $\| \bZ \| \leq \lambda$ is used to promote low-rank solutions $\bZ$~\cite{mc1} and the parameter $\lambda > 0$ is used to control its rank.
In practice, one can use the nuclear norm constraint $\| \bZ \|_* \leq \lambda$, which is a convex relaxation of the (non-convex) low-rank constraint $\rank(\bZ) \leq r$ \cite{mc1,svt}; 
alternatively, one can use the max-norm constraint $\| \bZ \|_\text{max}\leq \lambda$ (see \cite{squash} for the details). 
We select the only algorithm parameter $\lambda > 0$ via cross-validation.
We emphasize that this parameter-selection process of SPARFA-Lite is much more efficient than that for regular SPARFA, which has three algorithm parameters.

Since the gradient of the negative log-likelihood of the inverse logit link function can be computed efficiently, (P) can be solved efficiently via the FISTA framework \cite{fista}. Starting with an initialization of the matrix $\bZ$, at each inner iteration $\ell=1, 2, \ldots$, the algorithm performs a gradient step that aims at reducing the objective function $f(\bZ)$, followed by a projection step that makes the solution satisfy the constraint $\| \bZ \| \leq \lambda$. Both steps are repeated until convergence.

\sloppy

\sloppy

The \emph{gradient step} is given by
$\widehat{\bZ}^{\ell+1} \leftarrow \bZ^\ell - s_\ell \nabla f$,
where $s_\ell$ is the step-size at iteration $\ell$ (see \cite{qmc} for the details on step-size selection). 
The gradient of the objective function $f(\bZ)$ with respect to~$\bZ$ is given by
\begin{align*}  
[\nabla f]_{i,j} = \left \{ \begin{array}{ll}
\frac{\Phi'(L_{i,j}-Z_{i,j})-\Phi'(U_{i,j}-Z_{i,j})}{\Phi(U_{i,j}-Z_{i,j})-\Phi(L_{i,j}-Z_{i,j})}  &\text{if} \,\, (i,j) \in \Omega_\text{obs} \\[0.1cm]
0 & \text{otherwise},
\end{array} \right.
\end{align*} 
where the derivative of the inverse logit link function corresponds to $\Phi'(x) = \frac{1}{2+e^{-x}+e^x}$. 
The \mbox{$Q \times N$} matrices $\bU$ and $\bL$ contain the upper and lower bin boundaries corresponding to the measurements $Y_{i,j}$, i.e., we have $U_{i,j} = \omega_{Y_{i,j}}$ and $L_{i,j} = \omega_{Y_{i,j}-1}$.

The \emph{projection step} imposes low-rankness on $\bZ$. For the nuclear norm constraint case $\| \bZ \|_* \leq \lambda$, this step requires a projection onto the nuclear norm ball with radius $\lambda$, which can be performed by first computing the SVD of $\bZ$ followed by projecting the vector of singular values onto an $\ell_1$-norm ball with radius $\lambda$ (the details can be found in \cite{svt}). The resulting projection step corresponds to
\begin{align} \label{eq:shrinknuc} 
& \bZ^{\ell+1} \gets  \, \widetilde{\bU} \text{diag}(\vecs) \widetilde{\bV}^T, \,\,\,\text{with}\,\,\,\, \vecs = \text{P}_\lambda(\text{diag}(\bS)),
\end{align} 
where $\widetilde{\bU} \bS \widetilde{\bV}^T$ denotes the SVD of $\widehat{\bZ}^{\ell+1}$. The operator $\text{P}_\lambda(\cdot)$ denotes the projection of a vector onto the $\ellone$-norm ball with radius $\lambda$, which can be computed at low complexity \cite{qmc}.
For the max-norm constraint $\| \bZ \|_\text{max} \leq \lambda$, the projection step can be calculated efficiently by following the method put forward in \cite{squash}.
\sloppy
We emphasize that SPARFA-Lite is guaranteed to converge to a global optimum, since the problem $(\text{P})$ is convex.
\fussy

%

\section{SPARFA-Lite learning analytics} 
\label{sec:laca}

\sloppy
We now demonstrate how SPARFA-Lite can be used to perform LA. 
To this end, we assume that there is tag information available for each question, i.e., there are a set of $M$ user-defined labels (tags) associated with the $Q$ questions, with each question associated with at least one tag. 
We define the $Q\times M$ question-tag matrix $\bT$ with $T_{i,m} = 1$ if tag $m$ is associated to question $i$, and $T_{i,m} =0$ otherwise. 
We also define the $Q \times N$ matrix $\bA$ with $A_{i,j} = \Phi(Z_{i,j}) \in [0,1]$, which is the de-noised and completed version of the (partially observed) graded learner response matrix~$\bY$. 
Using both matrices $\bT$ and $\bA$, we can compute the $N \times M$ learner tag knowledge matrix $\bB$  with the entries $B_{j,m} = ({\sum_{i=1}^Q T_{i,m}})^{-1}{\widetilde{B}_{j,m}} \in [0,1]$, where $\widetilde{\bB} = \bA^T \bT$. 
The entries $B_{j,m}$ represent the de-noised concept knowledge of learner $j$ on tag $m$; large values represent good  knowledge of tag $m$, whereas small values represent poor tag knowledge.
This tag knowledge information is crucial for a PLS to perform LA.



\section{Experiments}
\label{sec:experiments}

We now compare SPARFA-Lite against existing factor analysis methods for predicting unobserved learner responses, using real-world educational datasets and demonstrate its efficacy in performing LA. 
All algorithm parameters are selected through cross-validation. 
All results are averaged over $25$ independent Monte--Carlo trials. 
\vspace{-0.2cm}

\subsection{Predicting unobserved learner responses}

\label{sec:real}

\sloppy

We first compare the performance of SPARFA-Lite in predicting unobserved graded learner responses with two state-of-the-art factor analysis algorithms.

\vspace{-0.5cm}
\paragraph{Datasets}
In this experiment, we use five different educational datasets for: 
(1) an undergraduate course on fundamentals of electrical engineering, consisting of $N=92$ learners answering $Q=203$ questions, with $99.5\%$ of the answers observed; 
(2) an undergraduate course on signals and systems, consisting of $N=41$ learners answering $Q=143$ questions, with $97.1\%$ of the answers observed;
(3) an undergraduate course on introduction to probability and statistics, consisting of $N=57$ learners answering $Q=107$ questions, with $68.9\%$ of the answers observed;
(4) a university entrance exam, consisting of $N=1706$ learners answering $Q=60$ questions, with $60.9\%$ of the answers observed;
and (5) another university entrance exam, consisting of $N=1564$ learners answering $Q=60$ questions, with $70.8\%$ of the answers observed.
The undergraduate course datasets are collected via OpenStax Tutor~\cite{ost}; see \cite{tesr} for the details on the university entrance exam dataset. 
Note that all of these datasets contain binary-valued graded learner responses, which is a special case of the general,  quantized model proposed above (with $P=2$ and $\{\omega_0, \omega_1, \omega_2\} = \{ -\infty, 0, \infty \}$).
For simplicity, we will refer to the individual datasets as Dataset~1-to-5, respectively.

\vspace{-0.4cm}
\paragraph{Experimental setup}
We now compare SPARFA-Lite against \mbox{CF-IRT}~\cite{logitfa} and SPARFA \cite{sparfa}, two established factor analysis methods that perform well in terms of predicting unobserved graded learner responses. 
To assess prediction performance on unobserved learner responses, we randomly puncture each dataset by removing $20\%$ of the observed learner responses in $\bY$ to form a test set. We then train all three algorithms on the rest of the observed learner responses and predict the unobserved responses in the test set. 
Since CF-IRT and SPARFA both have the number of concepts $K$ as a tuning parameter, we run both algorithms using a range of possible values of $K$ and select the value of $K$ that achieves the best prediction performance. 
For SPARFA-Lite, we only need to select the value of the single algorithm parameter $\lambda$ that controls $K$. 
To assess the prediction performance of all three algorithms, we use three well-established performance metrics: prediction accuracy (COR), prediction likelihood (LIK), and area under the receiver operation characteristic curve (AUC) \cite{kt}. The prediction accuracy corresponds to the percentage of correctly predicted responses. The prediction likelihood corresponds to the average the predicted likelihood of the unobserved responses, i.e., $\frac{\sum_{i,j:(i,j) \in \bar{\Omega}_\text{obs}} p(Y_{i,j} \mid Z_{i,j})}{|\bar{\Omega}_\text{obs}|}$, where $\bar{\Omega}_\text{obs}$ represents the set of learner responses in the test set. The area under curve is a commonly-used performance metric for binary classifiers (see \cite{kt} for the details). 

\fussy

\begin{table}
\vspace{-0.3cm}
\centering
\caption{Performance comparison of SPARFA-Lite vs. CF-IRT and SPARFA on predicting unobserved ratings for five  educational datasets. 
Bold numbers represent the best performance among the three algorithms. 
SPARFA-Lite achieves comparable performance to CF-IRT and SPARFA in all experiments and metrics at significantly lower computational complexity.}
\label{tbl:prediction}
\vspace{0.2cm}
\scalebox{.91}{
\begin{tabular}{lcccc}
\toprule
 &  & CF-IRT~\cite{logitfa} & SPARFA~\cite{sparfa} & {\bf SPARFA-Lite} \\
\midrule
\multirow{3}{*}{Dataset 1} & COR &0.8687  & 0.8711 &  \bf{0.8737}  \\
& LIK &    \bf{0.7286}  & 0.7195 & 0.7235  \\
& AUC &  0.8247  & 0.8056 & \bf{ 0.8299} \\
\midrule
\multirow{3}{*}{Dataset 2} & COR & 0.8061  & 0.8096  & \bf{0.8181}     \\
& LIK &   0.6393  &   \bf{0.6759}  & 0.6707       \\
& AUC &  0.7985  &   0.7285 & \bf{0.8047} \\
\midrule
\multirow{3}{*}{Dataset 3} & COR & \bf{0.7263}   &  0.7000    & 0.7200      \\
& LIK &   \bf{0.5876}   & 0.5334 & 0.5699    \\
& AUC &   \bf{0.7629} &  0.7116 & 0.7372 \\
\midrule
\multirow{3}{*}{Dataset 4} & COR & 0.6967   &0.7015  & \bf{0.7019}     \\
& LIK &   0.5538   &  \bf{0.5587}   & 0.5537  \\
& AUC &  0.7180 &   \bf{0.7249} &   0.7175   \\
\midrule
\multirow{3}{*}{Dataset 5} & COR & 0.6866  & 0.6880 & \bf{0.6903}     \\
& LIK &   0.5506  &    \bf{0.5536}  & 0.5505     \\
& AUC &   0.7457 &   \bf{0.7478} &  0.7472  \\
\bottomrule
\vspace{-0.4cm}
\end{tabular}
}
\vspace{-0.4cm}
\end{table}

\vspace{-0.4cm}
\sloppy
\paragraph{Results and discussion}
Table \ref{tbl:prediction} shows the mean of the performance metrics over 25 trials. We see that SPARFA-Lite achieves comparable performance as CF-IRT and SPARFA. 
Note that it outperforms CF-IRT and SPARFA on the most important performance metric--prediction accuracy (COR), with the exception of Dataset~3. 
%

We emphasize that SPARFA-Lite is computationally more efficient than CF-IRT and SPARFA, since it (i) has only a single algorithm parameter and  (ii) can be solved efficiently as it is a convex optimization problem. 
CF-IRT and SPARFA, in contrast, have multiple tuning parameters (including $K$) \cite{logitfa,sparfa}, which means one have to run them multiple times to conduct a grid search over all possible values of these parameters.
In particular, one Monte--Carlo trial of SPARFA-Lite on Dataset~1 only takes $3$\,sec, while CF-IRT and SPARFA require roughly $2$\,min.\ and $10$\,min.\ respectively, in MATLAB on a standard desktop PC with a $3.07$\,GHz Intel Core i7 processor (corresponding to $40\times$ and $200\times$ speed up). 
One can further reduce the computational complexity of SPARFA-Lite by replacing the nuclear norm constraint with the max-norm constraint~\cite{squash,1bitmax}.

\subsection{SPARFA-Lite learning analytics}
\label{sec:lacaexpt}

\vspace{-0.2cm}
\paragraph{Dataset and experimental setup}
In this experiment, we use data collected from a high-school algebra test conducted on Amazon's Mechanical Turk \cite{mechturkwebsite}. 
The dataset consists of the quantized (with $P=4$ ordinal values) graded responses of $N=99$ learners answering  $Q=34$ questions, and the learner responses are fully observed. 
A total of $M=13$ tags have manually been assigned to the questions. 
%
We use SPARFA-Lite to perform learning analytics on this dataset as described in \fref{sec:laca}.

\begin{table}
\vspace{-0.0cm}
\centering
\caption{Tag knowledge of selected learners. SPARFA-Lite performs robust LA by estimating each learner's tag knowledge from ordinal graded response data.} 
\vspace{-0.2cm}
\label{tbl:mturktag}
\scalebox{.8}{
    \begin{tabular}{lccc}
\toprule 
 & Simplifying  & \multirow{2}{*}{Geometry} & System of  \\ 
  &  expressions &  & equations \\ 
\midrule 
Class average &69 \% & 64\% & 30\%  \\
\midrule 
Best learner & 84\% & 79\% & 34\%  \\
Average learner & 70\% & 63\% & 24\%  \\
Worst learner & 32\% & 34\% & 43\%  \\
\bottomrule 
\end{tabular}}
\vspace{-0.4cm}
\end{table}

\vspace{-0.4cm}
\paragraph{Results and discussion}
Table~\ref{tbl:mturktag} shows the tag knowledge profile for a set of selected learners on the tags ``Simplifying expressions,'' ``Geometry,'' and ``Systems of equations.'' The first row of the table shows the mean tag knowledge of all learners (in precent), while rows 2--4 show the tag knowledge (in percent) for the best learner, an average learner, and the worst learner, respectively. 
Leveraging these tag knowledge profiles, a PLS can automatically provide personalized feedback to learners on their strengths and weaknesses, and automatically recommend learning resources for remedial studies. 
For example, for the average learner in Table~\ref{tbl:mturktag}, a PLS would alert them to focus on the tag ``System of equations'' and recommend them learning resources associated with this tag, because their tag knowledge is below the class average. 
Moreover, a PLS can use this analysis to provide feedback to course instructors on the average tag knowledge of the entire class, helping them to make timely adjustments to their future course plan.

\vspace{-0.2cm}
\section{Conclusions}
\sloppy
SPARFA-Lite is an efficient method that analyzes an incomplete set of quantized graded learner responses to questions to perform learning analytics. SPARFA-Lite achieves comparable or superior performance in predicting unobserved graded learner responses compared to existing factor-analysis methods, with significantly reduced computational complexity. 
\vspace{-0.2cm}







\bibliographystyle{abbrv}
\bibliography{sparfaclustbib}


\end{document}